\documentclass[manuscript,screen]{acmart}

\usepackage{times}
\usepackage{latexsym}
\usepackage{kotex}
\usepackage{url}
\usepackage{graphicx}
\usepackage{booktabs} 
\usepackage{multirow}
\usepackage[normalem]{ulem}
\useunder{\uline}{\ul}{}
\usepackage{enumitem}
\usepackage{makecell}
\setcellgapes{3.5pt}
\usepackage{stfloats}

\usepackage[T1]{fontenc}
\usepackage[utf8]{inputenc}
\usepackage{mathpazo}
\usepackage{textcomp}

\renewcommand{\textrightarrow}{$\rightarrow$}

\AtBeginDocument{%
  \providecommand\BibTeX{{%
    \normalfont B\kern-0.5em{\scshape i\kern-0.25em b}\kern-0.8em\TeX}}}

\begin{document}

\title{Speech Intention Understanding in a Head-final Language:\\A Disambiguation Utilizing Intonation-dependency}


\author{Won Ik Cho}
\affiliation{%
	\institution{Seoul National University, Dept. ECE and INMC}
}
\email{wicho@hi.snu.ac.kr}

\author{Hyeon Seung Lee}
\affiliation{%
	\institution{Seoul National University, Dept. ECE and INMC}
}
\email{hslee@hi.snu.ac.kr}

\author{Ji Won Yoon}
\affiliation{%
	\institution{Seoul National University, Dept. ECE and INMC}
}
\email{jwyoon@hi.snu.ac.kr}

\author{Seok Min Kim}
\affiliation{%
	\institution{Seoul National University, Dept. ECE and INMC}
}
\email{smkim@hi.snu.ac.kr}

\author{Nam Soo Kim}
\affiliation{%
	\institution{Seoul National University, Dept. ECE and INMC}
}
\email{nkim@snu.ac.kr}

\begin{abstract}
  For a large portion of real-life utterances, the intention cannot be solely decided by either their semantic or syntactic characteristics. Although not all the sociolinguistic and pragmatic information can be digitized, at least phonetic features are indispensable in understanding the spoken language. Especially in head-final languages such as Korean, sentence-final prosody has great importance in identifying the speaker's intention. This paper suggests a system which identifies the inherent intention of a spoken utterance given its transcript, in some cases using auxiliary acoustic features. The main point here is a separate distinction for cases where discrimination of intention requires an acoustic cue. Thus, the proposed classification system decides whether the given utterance is a fragment, statement, question, command, or a rhetorical question/command, utilizing the intonation-dependency coming from the head-finality. Based on an intuitive understanding of the Korean language that is engaged in the data annotation, we construct a network which identifies the intention of a speech, and validate its utility with the test sentences. The system, if combined with up-to-date speech recognizers, is expected to be flexibly inserted into various language understanding modules.
\end{abstract}




\maketitle

\section{Introduction}
Understanding the intention of a speech includes all aspects of phonetics, semantics, and syntax.
For example, even when an utterance is assigned with its syntactic structure of declarative/interrogative/imperative forms, the speech act may differ considering semantics and pragmatics \cite{stolcke2000dialogue}.
Besides, phonetic features such as prosody can influence the actual intention, which can be different from the illocutionary act that is grasped at first glance \cite{banuazizi1999real}.

In text data, punctuation plays a dominant role in conveying non-textual information.
However, in real life, where smart agents with microphones are widely used, critical phonetic features happen to be inadvertently omitted during transmission of data.
In many cases, a language model may punctuate a transcribed sentence from the speech recognition module. However, more accurate prediction on the intention is expected to be achieved by engaging in the acoustic data, as observed in the recent approach that co-utilizes audio and text \cite{gu2017speech}.

We were inspired by an idea that during this process, \textit{performing speech analysis for all the input can be costly for devices}. That is, devices can \textit{bypass the utterances whose intention is determined solely upon the text} and concentrate on acoustic analysis of the ambiguous ones. This approach might reduce the system malfunction from the speech with prosody that incurs confusion. Also, languages with low speech resource may benefit from this since a large portion of the utterances can be filtered as clear-cut ones via a text-based sieve.

In this study, the language of interest is Korean, a representative one with the head-final syntax.
Natural language processing on Korean is known to be burdensome; not only that the Korean language is agglutinative and morphologically rich, but also, the subject of a sentence sometimes happens to be omitted and be determined upon the context. Moreover, to make it challenging to understand the meaning only by text, the intention of certain types of sentences is significantly influenced by phonetic property of the sentence enders \cite{kim2005evidentiality}.
Consider the following sentence, of which the meaning depends on the sentence-final intonation:\medskip\\
(S1) 천천히 가고 있어

chen-chen-hi ka-ko iss-e

slowly go-PROG be-SE\footnote{Denotes the underspecified sentence enders; final particles whose role vary.}\medskip\\
With a high rise intonation, this sentence becomes a question (\textit{Are you/they going slowly?}), and given a fall or fall-rise intonation, becomes a statement (\textit{(I am) going slowly.}). 
Also, given a low rise or level intonation, the sentence becomes a command (\textit{Go slowly.}).
This phenomenon partially originates in the particular constituents of Korean utterances, such as multi-functional particle `-어 (-e)', or other sentence enders determining the sentence type \cite{pak2008types}. Although similar tendencies are observed in other languages as well (e.g., declarative questions in English \cite{gunlogson2002declarative}), the syntactical and morphological properties of the Korean language strengthen the ambiguity of the spoken utterances.

Here, we propose a partially improvable system that identifies the intention of a spoken Korean utterance, with disambiguation utilizing auxiliary acoustic information. The system categorizes input utterances into six categories of \textit{fragment, statement, question, command, and rhetorical question$\cdot$command}, regarding speech act. Although the system does not contain a speech recognition module, it receives the script (assumed correctly transcribed) and the acoustic feature (if required) of an utterance and infers the intention. A similar analysis based on parallel processing of audio and text has been widely utilized so far \cite{gu2017speech,gu2018hybrid}. However, ours accompanies the process of identifying and disambiguating intonation-dependent utterances, making the computation more efficient. To this end, total 61,225 lines of text utterances were semi-automatically collected or generated, including about 2K manually tagged lines. We claim the followings as our contribution:

\begin{itemize}[noitemsep]
	\item  A new kind of text annotation scheme that considers prosodic variability of the sentences (effective, but not restricted, to a head-final language), accompanying a freely-available corpus
	\item A two-stage system that consists of a text-based sieve and an audio-aided analyzer, which significantly reduces the computation/resource burden that exists in end-to-end acoustic systems
\end{itemize}

In the following section, we take a look at the literature of intention classification and demonstrate the background of our categorization. In Section 3 and 4, the architecture of the proposed system is described with a detailed implementation scheme. Afterward, the system is evaluated quantitatively and qualitatively with the validation and test set. Besides, we will briefly explain how our methodology can be adopted in other languages. 

\section{Background}

The most important among various areas related to this paper is the study on the sentence-level semantics. 
Unlike the syntactic concept of sentence form presented in \citet{sadock1985speech}, the speech act, or intention\footnote{In this paper, intention and act are often used interchangeably. In principle, the intention of an utterance is the object of grasping, and the act of speech is a property of the utterance itself. However, we denote determining the act of a speech, such as question and demand, as inferring the intention.}, has been studied in the area of pragmatics, especially illocutionary act and dialog act \cite{searle1976classification, stolcke2000dialogue}.
It is controversial how the speech act should be defined, but in this study, we refer to the previous works that suggest general linguistic and domain non-specific categorization. Mainly we concentrate on lessening vague intersections between the classes, which can be noticed between e.g., \textit{statement} and \textit{opinion} \cite{stolcke2000dialogue}, in the sense that some statements can be regarded as opinion and vice versa. Thus, starting from the clear boundaries between the sentence forms of \textit{declaratives, interrogatives}, and \textit{imperatives} \cite{sadock1985speech}, we first extended them to syntax-semantic level adopting discourse component (DC) \cite{portner2004semantics}. It involves \textit{common ground}, \textit{question set}, and \textit{to-do-list}: the constituent of the sentence types that comprise natural language. We interpreted them in terms of speech act, considering the obligations the sentence impose on the hearers; whether to answer (\textit{question}), to react (\textit{command}), or neither (\textit{statement}).

Building on the concept of reinterpreted discourse component, we took into account the rhetoricalness of question and command, which yielded additional categories of \textit{rhetorical question} (RQ) \cite{rohde2006rhetorical} and \textit{rhetorical command} (RC) \cite{kaufmann2016fine}. We claim that a single utterance falls into one of the five categories, if non-fragment and if given acoustic data\footnote{More elaborate definition of \textit{fragments} and intonation-dependency will be discussed in the next section.}. Our categorization is much simpler than the conventional DA tagging schemes \cite{stolcke2000dialogue,bunt2010towards}, and is rather close to tweet act \cite{vosoughi2016tweet} or situation entity types \cite{friedrich2016situation}. However, our methodology has strength in inferring the speech intention that less relies on the dialog history, which was possible by bringing into a new class of \textit{intonation-dependent utterances}, as will be described afterward.

At this point, it may be beneficial to point out that the term \textit{intention}  or \textit{speech act} is going to be used as a domain non-specific indicator of the utterance type. We argue that the terms are different from \textit{intent}, which is used as a specific action in the literature \cite{shimada2007case,liu2016attention,haghani2018audio}, along with the concept of \textit{item}, \textit{object} and \textit{argument}, generally for domain-specific tasks.
Also, unlike dialog management where a proper response is made based on the dialog history \cite{li2017dailydialog}, the proposed system aims to find the genuine intention of a single input utterance and recommend further action to the addressee.

We expect this research to be cross-lingually extended, but for data annotation, research on the Korean sentence types was essential.
Although the annotation process partly depended on the intuition of the annotators, we referred to the studies related to syntax-semantics and speech act of Korean \cite{han2000structure, pak2006jussive, seo2017syntax}, to handle some unclear cases regarding optatives, permissives, promisives, request$\cdot$suggestions, and rhetorical questions.

\section{System Concept}

The proposed system incorporates two modules as in Figure 1: (A) a module classifying the utterances into fragments, five clear-cut cases, and intonation-dependent utterances (FCI module), and (B) an audio-aided analyzer for disambiguation of the intonation-dependent utterances (3A module). Throughout this paper, \textit{text} refers to the sequence of symbols (letters) with the punctuation marks removed, which usually indicates the output of a speech recognition module. 
Also, \textit{sentence} and \textit{utterance} are interchangeably used to denote an input, but usually the latter implies an object with intention while the former does not necessarily.

\subsection{FCI module: Text-based sieving}

\textbf{Fragments (FR):} From a linguistic viewpoint, fragments often refer to single noun$\cdot$verb phrase where ellipsis occurred \cite{merchant2005fragments}. However, in this study, we also include some incomplete sentences whose intention is underspecified. If the input sentence is not a fragment, then it is assumed to belong to the clear-cut cases or be an intonation-dependent utterance\footnote{There were also context-dependent cases where the intention was hard to decide even given intonation, but the portion was tiny.}. 

\begin{figure}
	\centering
	\includegraphics[width=0.6\columnwidth]{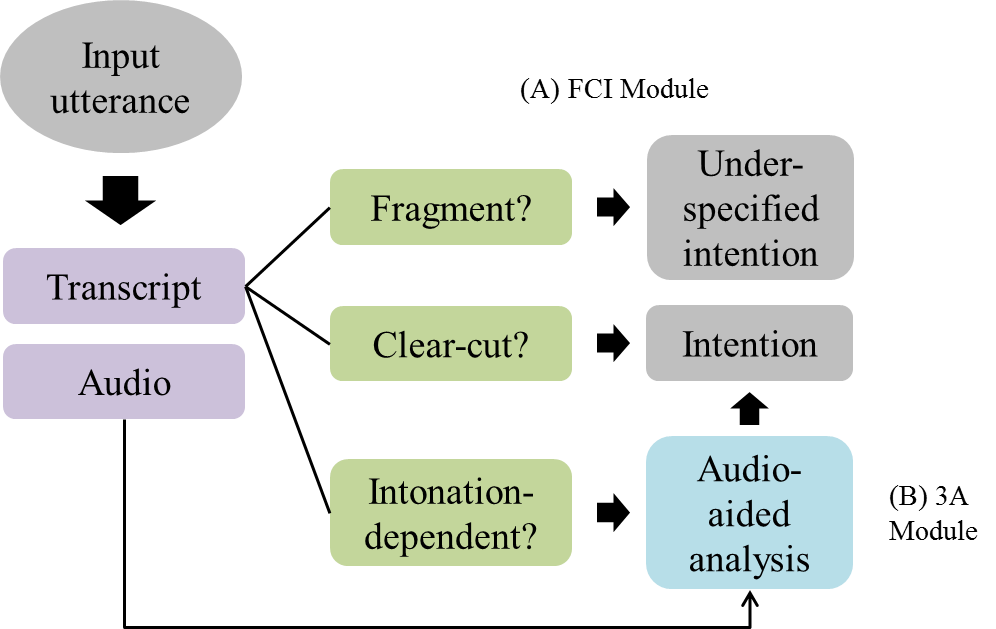}
	\caption{A brief illustration on the structure of the proposed system.
	} \label{fig:fig2}
\end{figure} 

It is arguable that fragments are interpreted as \textit{command} or \textit{question} under some circumstances. However, we found it difficult to assign a specific intention to them even given audio, since it mostly requires the dialog history which we do not handle within this study. We observed that a large portion of the intention concerning context is represented in the prosody, which led us to define prosody-sensitive cases afterward separately.  Whereas for fragments, discerning such implication is not feasible in many cases. Therefore, we decided to leave the intention of the fragments underspecified, and let them be combatted with the help of the context in the real world usage.\medskip\\
\textbf{Clear-cut cases (CCs):} Clear-cut cases incorporate the utterances of the five categories: \textit{statement, question, command, rhetorical question}, and \textit{rhetorical command}, as described detailed in the annotation guideline\footnote{Will be published online.} with the examples.
Briefly, questions are the utterances that require the addressee to answer, and commands are the ones that require the addressee to act.
Even if sentence form is declarative, words such as \textit{wonder} or \textit{should} can make the sentence question or command. Statements are descriptive and expressive sentences that do not apply to both cases. 

Rhetorical questions (RQ) are the questions that do not require an answer because the answer is already in the speaker's mind \cite{rohde2006rhetorical}. Similarly, rhetorical commands (RC) are idiomatic expressions in which imperative structure does not convey a to-do-list that is mandatory (e.g., \textit{Have a nice day}) \cite{han2000structure,kaufmann2016fine}.
The sentences in these categories are functionally similar to statements but were categorized separately since they usually show non-neutral tone.

\begin{table}[]
	\centering
	\resizebox{0.6\columnwidth}{!}{%
		\begin{tabular}{|c|c|c|c|}
			\hline
			\textit{\textbf{\begin{tabular}[c]{@{}c@{}}Discourse component\\ / Sentence form\end{tabular}}} & \textbf{\begin{tabular}[c]{@{}c@{}}Common \\ Ground\end{tabular}} & \textbf{Question Set} & \textbf{To-do List} \\ \hline
			\textbf{Declaratives} & \textit{\begin{tabular}[c]{@{}c@{}}Statements, \\ RQ, RC\end{tabular}} & \textit{Question} & \textit{Command} \\ \hline
			\textbf{Interrogatives} & \textit{RQ} & \textit{Question} & \textit{Command} \\ \hline
			\textbf{Imperatives} & \textit{RC} & \textit{Question} & \textit{Command} \\ \hline
		\end{tabular}%
	}
	\caption{A simplified annotation scheme regarding discourse component and sentence form: Discourse component (DC) in the table implies the concept that extends the original formal semantic property \cite{portner2004semantics} to speech act level.}
	\label{my-label}
\end{table}

In making up the guideline, we carefully looked into the dataset so that the annotating schemes can cover the ambiguous cases. As stated in the previous section, we referred to \citet{portner2004semantics} to borrow the concept of discourse component and extended the formal semantic property to the level of pragmatics. That is, we searched for a question set (QS) or to-do-list (TDL) which makes an utterance a valid directive in terms of speech act \cite{searle1976classification}, taking into account non-canonical and conversation-style sentences which contain idiomatic expressions and jargons. We provide a simplified criterion with Table 1. \medskip\\
\textbf{Intonation-dependent utterances (IU):} With decision criteria for the clear-cut cases, we investigated \textit{whether the intention of a given sentence can be determined with punctuations removed}. The decision process requires attention to the final particle and the content. 
There has been a study on Korean sentences which handled with final particles and adverbs \cite{nam2014novel}. However, up to our knowledge, there has been no explicit guideline on a text-based identification of the utterances whose intention is influenced by intonation. Thus, we set up some principles, or the rules of thumb, concerning the annotating process. Note that the last two are closely related with the maxims of conversation \cite{levinson2000presumptive}, e.g., \textit{``Do not say more than is required."} or \textit{``What is generally said is stereotypically and specifically exemplified."}.
\begin{itemize}[noitemsep]
	\item Consider mainly the sentence-final intonation, since sentence-middle intonation usually concerns the topic. Moreover, the sentence-middle intonation eventually influences the sentence-final one in general\footnote{Except some case regarding \textit{wh-} particles.}.
	\item Consider the case where a \textit{wh-}particle is interpreted as an existential quantifier (\textit{wh-} intervention). 
	\item If the subject is missing, assign all the agents (1\textsuperscript{st} to 3\textsuperscript{rd} person) and exclude the awkward ones.
	\item Note the presence of vocatives.
	\item Avoid the loss of felicity that is induced if adverbs or numerics are used in a question.
	\item Do not count the sentences containing excessively specific information as questions. If so, some sentences can be treated as declarative questions in an unnatural way.
\end{itemize}

\subsection{3A module: Disambiguation via acoustic data}

To perform an audio-aided analysis, we constructed a multi-modal\footnote{To be strict, the system takes a speech as an only input and makes a decision assuming perfect transcription, that it is not fully \textit{multi-}modal; rather, it might have to be called as a co-utilization of audio and text. However, for simplicity, we denote the approach above as a multi-modal one.} network as proposed in \citet{gu2017speech} (Figure 2). The paper suggests that the multi-modal network shows much higher performance compared with \textit{only-speech} and \textit{only-text} systems, which will be investigated in this study as well. In our model, the multi-modal network only combats the intonation-dependent utterances, while in the previous work, it deals with all the utterances tagged with intention. The two cases, utilizing the multi-modal network as a submodule or as a whole system, are to be examined via comparison.

\begin{figure}
	\centering
	\includegraphics[width=0.6\columnwidth]{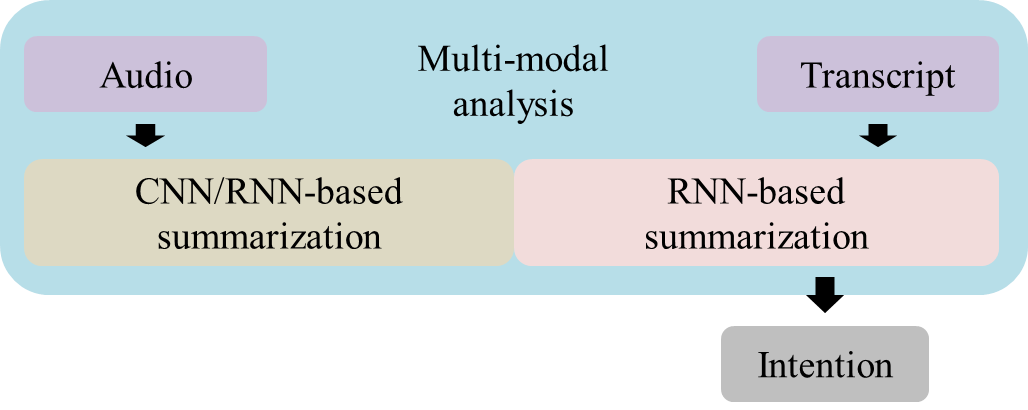}
	\caption{A brief illustration on the structure of the 3A module that adopts multi-modal network.
	} \label{fig:fig2}
\end{figure} 

For \textbf{acoustic feature}, mel spectrogram (MS) was obtained by mapping a magnitude spectrogram onto mel basis. From each audio of length $L_f$ denoting the number of frames, we extracted MS of size ($L_f$, $N$) where $N$ indicates the number of frequency bins. Here the length of the fast Fourier transform (FFT) window and hop length was fixed to 2,048 and 512 respectively. Since all the utterances differ in length and the task mainly requires the tail of a speech, we utilized the last 300 frames, with zero-padding for the shorter utterances. $N$ was set to 128. To deal with the syllable-timedness of Korean, MS was augmented with the energy contour that emphasizes the syllable onsets, making up a final feature (MS + E) of size (300, 129). We maintain identical feature engineering for the speech analysis throughout this study. For the training and inference phase, CNN/RNN-based summarization is utilized based on an experiment.

For \textbf{textual feature}, which is co-utilized with the acoustic feature in the multi-modal network and is solely used in the FCI module, a character-level encoding is utilized so as to avoid the usage of a morphological analyzer. In our pilot study adopting the morphological decomposition, the performance was shown lower in general, possibly due to the train and test utterances being colloquial. Thus, by using the raw text form as an input, we prevent the malfunction of an analyzer which may not fit with the conversation-style and non-canonical sentences. Also, this might help guarantee that our experiment, which adopts only the perfect transcript, have a similar result with the case that adopts the ASR transcription\footnote{ASR errors in Korean usually regard character-level mismatches, where characters approximately correspond to subwords in other languages.}.
To this end, we adopted recently distributed pre-trained Korean character vectors \cite{cho2018real} which were obtained based on skip-gram \cite{mikolov2013distributed} of fastText \cite{bojanowski2016enriching}. The vectors are 100-dim dense, encompassing the distributive semantics over the scripted corpus. Similar to the case of the acoustic feature, the character sequence is padded from the end of the sentence, and spaces are counted as a character to provide the segmentation information. We counted the last 50 characters, making up a feature of size (50, 100). For the training, CNN or RNN-based summarization is used as in the audio, upon performance. The chosen model is to be demonstrated in the following section, accompanied by a comparison with the other models.

\begin{table}[]
	\centering
	\makegapedcells
	\resizebox{0.6\columnwidth}{!}{%
		\begin{tabular}{@{}ccccc@{}}
			\toprule
			\multirow{2}{*}{\textbf{\begin{tabular}[c]{@{}c@{}}Text-based sieving\\ (for FCI module)\end{tabular}}} & \multirow{2}{*}{\textbf{Intention}} & \multicolumn{3}{c}{\textbf{Instances}} \\ \cmidrule(l){3-5} 
			&  & \textbf{Corpus 20K} & \textbf{Text} & \textbf{Speech} \\ \midrule
			\textit{\textbf{Fragments}} & - & 384 & 6,009 & 145 \\ \midrule
			\multirow{5}{*}{\textit{\textbf{Clear-cut cases}}} & statements & 8,032 & 18,300 & 4,227 \\
			& questions & 3,563 & 17,869 & 636 \\
			& commands & 4,571 & 12,968 & 879 \\
			& RQs & 613 & 1,745 & 954 \\
			& RCs & 572 & 1,087 & 159 \\ \midrule
			\textit{\textbf{\begin{tabular}[c]{@{}c@{}}Intonation-dependent\\ utterances\end{tabular}}} & \begin{tabular}[c]{@{}c@{}}unknown\\ (for 3A module)\end{tabular} & 1,583 & 3,277 & 440* \\ \midrule
			& \textbf{Total} & \textbf{19,318} & \textbf{61,255} & \textbf{7,000} \\ \bottomrule
		\end{tabular}%
	}
	\caption{Composition of the corpora: First two corpora are text-only, and the last one incorporates speech. IUs(*) for the speech corpus overlap with the FRs and CCs.}
	\label{my-label}
\end{table}

\section{Experiment}

\subsection{Corpora}

To cover a variety of topics, the utterances used for training and validation of the FCI module were collected from (i) the corpus provided by 
Seoul National University Speech Language Processing Lab\footnote{http://slp.snu.ac.kr/},
(ii) the set of frequently used words, released from the National Institute of Korean Language\footnote{https://www.korean.go.kr/}, and (iii) manually created questions/commands.

From (i) which contains short utterances with topics covering e-mail, housework, weather, transportation, and stock, 20K lines were randomly selected, and three Seoul Korean L1 speakers classified them into the seven categories of FR, IU, and five CC cases (Table 2, Corpus 20K). Annotators were well informed on the guideline and had enough debate on the conflicts that occurred during the annotating process. The resulting inter-annotator agreement (IAA) was $\kappa$ = 0.85 \cite{fleiss1971measuring} and the final decision was done by majority voting.

Taking into account the shortage of the utterances, (i)-(iii) were utilized in the data supplementation. (i) contains various neutral statements and the rhetorical utterances. In (ii), the single nouns were collected and augmented to FR. The utterances in (iii) which were generated with purpose, were augmented to \textit{question} and \textit{command} directly. The composition of the final corpus is stated in Table 2. 

For training of the multi-modal system, we annotated a speech corpus of size 7,000 utilized in \citet{lee2018acoustic}. The annotation process follows the guideline introduced in Section 3.1; due to the tagging being relatively straightforward with the help of the acoustic information, the corpus was annotated by one of the annotators. The composition of the speech corpus is also presented in the table.

\subsection{Implementation}

The system architecture can be described as a combination of convolutional neural network (CNN) \cite{krizhevsky2012imagenet,kim2014convolutional} and self-attentive bidirectional long short-term memory (BiLSTM-Att) \cite{schuster1997bidirectional,lin2017structured}. For CNN, five convolution layers were stacked with the max-pooling layers in between, summarizing the distributional information lying in the input of a spectrogram (acoustic features) or a character vector sequence (textual features, although used for CNN only in the pilot study). For BiLSTM, the hidden layer of a specific timestep was fed together with the input of the next timestep, to infer the subsequent hidden layer in an autoregressive manner. For a self-attentive embedding, the context vector whose length equals to that of the hidden layers of BiLSTM, was jointly trained along with the network so that it can provide the weight assigned to each hidden layer. The input format of BiLSTM equals to that of CNN except for the channel number which was set to 1 (single channel) in the CNN models. The architecture specification is provided in Table 3.

\begin{table}[]
	\centering
	\resizebox{0.65\columnwidth}{!}{%
		\begin{tabular}{|c|c|c|}
			\hline
			& \multicolumn{2}{c|}{\textbf{Specification}} \\ \hline
			\multirow{3}{*}{\textbf{\begin{tabular}[c]{@{}c@{}}CNN\\ (audio)\end{tabular}}} & \begin{tabular}[c]{@{}c@{}}Input size\\ (single channel)\end{tabular} & (\textit{$L_{f}$}, 129, 1) \\ \cline{2-3} 
			& \# Conv layer & 5 \\ \cline{2-3} 
			& \begin{tabular}[c]{@{}c@{}}Window size (\# filters)\\ (→ Batch normalization)\\ → Max pooling size\\ (→ Dropout)\end{tabular} & \begin{tabular}[c]{@{}c@{}}5 by 5 (32) → 2 by 2\\ 5 by 5 (64) → 2 by 2\\ 3 by 3 (128) → 2 by 2\\ 3 by 3 (32) → 2 by 1\\ 3 by 3 (32 → 2 by 1\end{tabular} \\ \hline
			\multirow{3}{*}{\textbf{\begin{tabular}[c]{@{}c@{}}CNN\\ (text)\end{tabular}}} & \begin{tabular}[c]{@{}c@{}}Input size\\ (single channel)\end{tabular} & (\textit{$L_{max}$}, 100, 1) \\ \cline{2-3} 
			& \# Conv layer & 2 \\ \cline{2-3} 
			& \begin{tabular}[c]{@{}c@{}}Window size (\# filters)\\ → Max pooling size\end{tabular} & \begin{tabular}[c]{@{}c@{}}3 by 100 (32) → 2 by 1\\ 3 by 1 (no max-pooling)\end{tabular} \\ \hline
			\multirow{3}{*}{\textbf{\begin{tabular}[c]{@{}c@{}}BiLSTM\\ -Att\end{tabular}}} & Input size & \begin{tabular}[c]{@{}c@{}}( \textit{$L_{f}$}, 129) (audio)\\ ( \textit{$L_{max}$}, 100, 1) (text)\end{tabular} \\ \cline{2-3} 
			& Hidden layer nodes & 128 (64 x 2) \\ \cline{2-3} 
			& Context vector size & 64 \\ \hline
			\textbf{MLP} & Hidden layer nodes & 64 or 128 \\ \hline
			\multirow{4}{*}{\textbf{Others}} & Optimizer & \begin{tabular}[c]{@{}c@{}}Adam (0.0005)\\  \cite{kingma2014adam}\end{tabular}  \\ \cline{2-3} 
			& Batch size & 16 \\ \cline{2-3} 
			& Dropout & 0.3 (for CNN/MLP) \\ \cline{2-3} 
			& Activation & \begin{tabular}[c]{@{}c@{}}ReLU (CNN/MLP)\\ Softmax (attention, output)\end{tabular} \\ \hline
		\end{tabular}%
	}
	\caption{Architecture specification. \textit{$L_{f}$} (for the audio frames) was set to 300 and \textit{$L_{max}$} (for the character sequence) was set to 50, considering the utterances' length. Taking into account the syntactic property of the Korean language, sentence-final frames/syllables were utilized. The batch normalization \cite{ioffe2015batch} and dropout \cite{srivastava2014dropout} were utilized only for the CNN (audio) and the MLPs.}
	\label{my-label}
\end{table}

First, (A) \textbf{FCI module} was constructed using a character BiLSTM-Att \cite{lin2017structured} alone, which shows the best performance among the implemented models (Table 4). CNN is good at recognizing a syntactic distinction that comes from the length of utterances or the presence of specific sentence enders, but not appropriate for handling the scrambling of the Korean language, worsening the performance in the concatenated network. The structure of the FCI module equals to that of the \textit{only-text} model, which is implemented in Section 4.4 for the comparison.

\begin{table}[]
	\centering
	\resizebox{0.53\columnwidth}{!}{%
		\begin{tabular}{|c|c|c|}
			\hline
			\textbf{Models} & F1 score & accuracy \\ \hline
			charCNN & 0.7691 & 0.8706 \\ \hline
			charBiLSTM & 0.7811 & 0.8807 \\ \hline
			charCNN + charBiLSTM & 0.7700 & 0.8745 \\ \hline
			charBiLSTM-Att & \textbf{0.7977} & \textbf{0.8869} \\ \hline
			charCNN + charBiLSTM-Att & 0.7822 & 0.8746 \\ \hline
		\end{tabular}%
	}
	\caption{Model performance for FCI module.}
	\label{my-label}
\end{table}

Next, for (B) \textbf{3A module}, especially in abstracting the acoustic features, the concatenation of CNN and BiLSTM-Att was utilized, in the sense that prosody concerns both shape-related properties (e.g., mel spectrogram) and sequential information. Also, as expected, the models which use root mean square energy (RMSE) sequence seem to emphasize the syllable onsets that mainly affect the pitch contour in Korean.
For the textual features, a character BiLSTM-Att is adopted as in the FCI module. Eventually, the output layer of the acoustic feature is concatenated with the output layer of the character BiLSTM-Att, making up a thought vector that concerns both audio and text. The concatenated vector is fed as an input of an MLP to infer the final intention. The structure of the 3A module equals to that of the multi-modal network, and partially to the \textit{only-speech} model which is implemented for the evaluation.

In brief, (A) \textbf{FCI module} adopts a self attentive char-BiLSTM (acc: 0.88, F1: 0.79). 
For (B) \textbf{3A module}, the networks each utilizing audio (CNN and BiLSTM-Att merged) and text (char BiLSTM-Att) were jointly trained via simple concatenation, to make up a multi-modal network (acc: 0.75, F1: 0.61). 
For all the modules, the dataset was split into train and validation set with the ratio of 9:1. The class weight was taken into account in the training session concerning the imbalance of the volume for each utterance type. The implementation of the whole system was done with Librosa\footnote{https://github.com/librosa/librosa}, fastText\footnote{https://pypi.org/project/fasttext/} and Keras \cite{chollet2015keras}, which were used for extracting acoustic features, embedding character vectors, and making neural network models, respectively. 

\subsection{Result}

To interpret the performance, we analyzed the validation result for each module quantitatively and qualitatively.

The \textbf{FCI module} receives only a text input and classifies it into seven categories; thus, we made up a  confusion matrix with the validation result (Table 5). Note that  FRs, statements, questions, and commands show high accuracy ($>$ 85\%) while others show lower ($<$ 65\%). RQs show the lowest accuracy (48\%) and a large portion of the wrong answers were related to the utterances that are even difficult for a human to disambiguate since nuance is involved. It was encouraging that the frequency of false alarms regarding RCs and IUs is low in general. For RCs, the false alarms might induce an excessive movement of the addressee (e.g., AI agents) and for IUs, an unnecessary analysis on the speech data could have been performed.

\begin{table}[]
	\centering
	\resizebox{0.55\columnwidth}{!}{%
		\begin{tabular}{|c|c|c|c|c|c|c|c|}
			\hline
			\textbf{Pred \textbackslash Ans} & \textit{\textbf{FR}} & \textit{\textbf{S}} & \textit{\textbf{Q}} & \textit{\textbf{C}} & \textit{\textbf{RQ}} & \textit{\textbf{RC}} & \textit{\textbf{IU}} \\ \hline
			\textit{\textbf{Fragment}} & 586 & 6 & 5 & 3 & 2 & 2 & 7 \\ \hline
			\textit{\textbf{Statement}} & 6 & 1,692 & 34 & 111 & 44 & 15 & 81 \\ \hline
			\textit{\textbf{Question}} & 0 & 30 & 1,720 & 44 & 32 & 2 & 17 \\ \hline
			\textit{\textbf{Command}} & 6 & 4 & 33 & 1,090 & 4 & 19 & 20 \\ \hline
			\textit{\textbf{Rhetorical Q}} & 1 & 7 & 12 & 6 & 82 & 1 & 0 \\ \hline
			\textit{\textbf{Rhetorical C}} & 0 & 12 & 1 & 11 & 1 & 68 & 1 \\ \hline
			\textit{\textbf{Into-dep U}} & 2 & 42 & 10 & 15 & 7 & 1 & 194 \\ \hline
		\end{tabular}%
	}
	\caption{Confusion matrix for the validation of the FCI module.}
	\label{my-label}
\end{table}

\begin{table}[]
	\centering
	\resizebox{0.45\columnwidth}{!}{%
		\begin{tabular}{|c|c|c|c|c|c|c|}
			\hline
			\textbf{Pred \textbackslash Ans} & \textit{\textbf{FR}} & \textit{\textbf{S}} & \textit{\textbf{Q}} & \textit{\textbf{C}} & \textit{\textbf{RQ}} & \textit{\textbf{RC}} \\ \hline
			\textit{\textbf{Fragment}} & 13 & 3 & 0 & 0 & 1 & 0 \\ \hline
			\textit{\textbf{Statement}} & 4 & 363 & 11 & 22 & 28 & 4 \\ \hline
			\textit{\textbf{Question}} & 0 & 7 & 43 & 2 & 6 & 1 \\ \hline
			\textit{\textbf{Command}} & 2 & 24 & 7 & 43 & 3 & 6 \\ \hline
			\textit{\textbf{Rhetorical Q}} & 1 & 14 & 16 & 1 & 64 & 1 \\ \hline
			\textit{\textbf{Rhetorical C}} & 0 & 0 & 3 & 3 & 0 & 4 \\ \hline
		\end{tabular}%
	}
	\caption{Confusion matrix for the validation of the multi-modal network. IUs were all disambiguated.}
	\label{my-label}
\end{table}

There were some wrong answers regarding the prediction as statements. We found that most of them show a long sentence length that can confuse the system with the descriptive expression; especially among the ones which are originally question, command, or RQ. For example, some of the misclassified commands contained a modal phrase (e.g., -야 한다 (\textit{-ya hanta}, should)) that is frequently used in prohibition or requirement. This let the utterance be recognized as a descriptive one. Also, we could find some errors incurred by the morphological ambiguity of Korean. For example, `베란다 (\textit{peylanta}, a terrace)' was classified as a statement due to the existence of `란다 (\textit{lanta}, declarative sentence ender)', albeit the word (a single noun) has nothing to do with descriptiveness. 

For the \textbf{3A module}, a confusion matrix was constructed considering the multi-modal network which yields one of the six labels (after excluding IU) (Table 6) as an output. Statements showed high accuracy (88\%), as it is assumed that much of the inference is affected by the textual analysis. For the others, due to the scarcity of the sentence type (FR, RC) or the identification being challenging (Q and RQ), the performance was not encouraging. However, it is notable that the accuracy regarding FR (65\%) and RQ (62\%) is much higher than the rest. It seems that the distinct acoustic property such as the length of speech or the fluctuation of magnitude might have positively affected the inference. This is the point where the audio-aided analysis displays its strength, as shown in the performance gap between the RQs of the FCI module (48\%) \footnote{Although the fairness of the comparison is not guaranteed since the adopted corpus is different, the tendency is still appreciable.}. 

For the utterances with less distinct acoustic property, such as commands (59\%) and RCs ($<$ 30\%), a significant amount of misclassification was shown, mostly being identified as a statement. Despite the noticeable phonetic property such as rising intonation, the lower performance was achieved with the pure questions (53\%). It seems to originate in the dominant volume of the RQs in the dataset, which regards the corpus being scripted.

Overall, we found out that the utilization of acoustic features helps understand the utterances with distinct phonetic properties. However, it might deter the correct inference between the utterance types that are not, especially if the volume is small or the sentence form is similar.

\subsection{Evaluation}

To investigate the practical reliability of the whole system, we separately constructed a test set of size 2,000. Half of the set contains 1,000 challenging utterances. It consists of drama lines with the punctuation marks removed, recorded audio \cite{lee2018acoustic}, and manually tagged six-class label (FR and five CCs). For another half, 1,000 sentences in the corpus (i) (not overlapping with the ones randomly chosen for FCI module) were recorded and manually tagged. The former incorporates highly scripted lines while the latter encompasses the utterances in real-life situations. By binding them, we assign fairness to the test of the models trained with both types of data. 
Here, we compare four systems in total: (a) \textbf{only-speech}, (b) \textbf{only-text}, (c) \textbf{multi-modal}\footnote{We’re aware that our module is incomplete for `multi-modal', since ours adopts script, not ASR result. However, to minimize the gap between script-given model and real-world applications, we used the character-level embeddings, which are robust to ASR errors.} \cite{gu2017speech}, and (d) \textbf{text + multi-modal}.

For (a), we adopted the speech corpus annotated in Section 4.1.  All the IUs were tagged with their genuine intention regarding audio. For (b), we removed IUs from the corpora (labeled as \textit{-IU} in the table) and performed 6-class classification. For a fair comparison with the speech-related modules (a, c), we utilized the corpus of small size (7,000)\footnote{The model trained with large-scale corpus will be introduced in a while.}. For (c), we utilized both script and speech. Note that (c) equals to the structure suggested in (B) (Figure 2). 
For (d), the proposed system, a cascade structure of the FCI module and the multi-modal network was constructed.

The overall dataset size, training scheme, neural network architecture, computation, and evaluation result are described in Table 7. Notably, the model with a text-based sieve (d) yields a comparable result with the multi-modal model (c) while reducing the computation time to about 1/20. The utility of the text-based sieve is also observed in the performance of (b), which is much higher than (a) and close to (c, d).

With the large-scale corpus constructed in Section 4.1, the models which show much higher performance were obtained (b\textsuperscript{+}, d\textsuperscript{+}). The models were enhanced with both accuracy and F1 score by a large portion compared to the models trained with the small corpus while preserving the short inference time. This kind of advance seems to be quite tolerable, considering that many recent breakthroughs in NLU tasks accompanied pretrained language inference systems that take benefit from out-of-data information.

Beyond our expectation, the model which utilizes only the text data (b\textsuperscript{+}) showed the higher performance, both in accuracy and F1-score. This does not necessarily contradict our hypothesis since we did not assume that hybrid or multi-modal system is better than \textit{only-speech} or \textit{only-text} model. Instead, we interpret this result as the advantage of aggregating the concept of prosodic ambiguity into the corpus construction. The result was achieved by making up a corpus, which reduces the intervention of ambiguity (61K -IU), in a less explored head-final language. It mainly concerns disentangling the vagueness coming from some sentence enders such as `-어 (-e)' or `-지 (-ci)', and letting the analysis concentrate on content and nuance.

\begin{table*}[]
	\centering
	\makegapedcells
	\resizebox{\textwidth}{!}{%
		\begin{tabular}{|c|c|c|c|c|c|c|c|}
			\hline
			\textbf{Model} & \textbf{Symbol} & \textbf{Speech corpus} & \textbf{Text corpus} & \textbf{Training scheme} & \textbf{Architecture} & \multicolumn{1}{l|}{\textbf{Computation}} & \textbf{Accuracy (F1-score)} \\ \hline
			\textit{\textbf{Only-speech}} & (a) & o (7K) & x & end2end & CNN + RNN & 3m 20s & 50.00\% (0.1972) \\ \hline
			\textit{\textbf{Only-text}} & (b) & x & o (7K -IU) & end2end & Char RNN & 8s & 57.20\% (0.3474) \\ \hline
			\textit{\textbf{Multi-modal}} & (c) & o (7K) & o (7K) & end2end & (a) + (b) & 3m 30s & 58.65\% (0.3706) \\ \hline
			\textit{\textbf{\begin{tabular}[c]{@{}c@{}}Text\\ + Multi-modal\end{tabular}}} & (d) & o (7K) & o (7K) & 2-stage & (b) → (c) & 10s & 58.50\% (0.3814) \\ \hline
			\textit{\textbf{Text-only (large)}} & (b\textsuperscript{+}) & x & o (61K -IU) & end2end & Char RNN & 8s & 75.65\% (0.5627) \\ \hline
			\textit{\textbf{\begin{tabular}[c]{@{}c@{}}Text  (large)\\ + Multi-modal\end{tabular}}} & (d\textsuperscript{+}) & o (7K) & o (61K) & 2-stage & (b\textsuperscript{+}) → (c) & 10s & 75.55\% (0.5227) \\ \hline
		\end{tabular}%
	}
	\caption{Specification of the models compared in the evaluation. (a-c) denote the previous approaches, and (d) indicates the proposed two-stage-system, where all (a-d) are trained/tested with 7K (small-size) utterances. In text corpora, \textit{-IU} implies the omission of IUs. In training scheme, end2end means that the whole network is trained at once. In architecture, RNN denotes BiLSTM-Att. The computation denotes the time spent in the inference of 1,000 utterances.}
	\label{my-label}
\end{table*}

\subsection{Discussion}

Given 7K speech-script pairs for train/test in (a-d), we interpret (b\textsuperscript{+})/(d\textsuperscript{+}) as also the proposed systems\footnote{(c+) was not available since the resource is limited, and bringing additional speech dataset for the evaluation was not aimed in this study.}. They utilize the larger dataset (61K), that is easier in annotation/acquisition than speech, to boost the performance regarding low-resource speech data. Despite some weak points of the FCI/3A module, the whole proposed systems (d, b\textsuperscript{+}, d\textsuperscript{+}) illustrate a useful methodology for a text intention disambiguation utilizing the potential acoustic cue. 

Besides, we want to clarify some points about the head-final language and our work’s scalability. Head-final syntax regards languages such as Japanese/Korean/Tamil (considering only the rigid head-final ones). We used the term ‘a head-final language’ since we have executed the experiment only for the Korean language. However, we claim that the scheme can be expanded to the other languages that display underspecified sentence enders or \textit{wh-} particles \textit{in-situ}. Moreover, we expect the scheme to be adopted to non-head-final languages that incorporate the type of utterances whose intention is ambiguous without prosody/punctuation (e.g., declarative questions in English). 

Referring again to the literature, the result can be compared to the case of utilizing a fully multi-modal system as suggested for English \cite{gu2017speech}, where the accuracy of 0.83 was obtained with the test set split from the English corpus. Such kind of systems are uncomplicated to construct and might be more reliable in the sense that less human factors are engaged in the implementation. Nevertheless, our approach is meaningful for the cases where there is a lack of resource in labeled speech data. The whole system can be partially improved by augmenting additional text or speech. Also, the efficiency of the proposed system lies in utilizing the acoustic data only for the text that requires additional prosodic information. Resultingly, it lets us avoid redundant computation and prevent confusion from unexpected prosody of users. 

We do not claim that our approach is optimum for intention identification. However, we believe that the proposed scheme might be helpful in the analysis of some low-resource languages since text data is easier to acquire and annotate. We mainly target the utilization of our approach in goal-oriented/spoken-language-based AI systems, where the computation issue is challenging to apply acoustic analysis for all the input speech.

\section{Conclusion}

In this paper, we proposed a text-based sieving system for the identification of speech intention. The system first checks if input speech is a fragment or has a confidently determinable act. If neither, it conducts an audio-aided decision, associating the underspecified utterance with the inherent intention. For a data-driven training of the modules, 7K speech and 61K text data were collected or manually tagged, with a fairly high inter-annotator agreement. The proposed NN-based systems yield a comparable result with or outperform the conventional approaches with an additionally constructed test set. 

Our goal in theoretical linguistics lies in making up a new speech act categorization that aggregates potential acoustic cue. It was shown to be successful and is to be discussed more thoroughly via a separate article. However, more importantly, a promising application of the proposed system regards SLU modules of smart agents, especially the ones targeting free-style conversation with a human. 
It is why our speech act categorization includes directiveness and rhetoricalness of an utterance. 
Using such categorization in dialog managing might help the people who are not familiar with the way to talk to intelligent agents.
It widens the accessibility of the speech-driven AI services and also sheds light to the flexible dialog management.

Our future work aims to make the sieve more rigorous and to augment a precise multi-modal system for 3A module, which can be reliable by making up an elaborately tagged speech DB. Besides, for a real-life application, the analysis that can compensate the ASR errors will be executed. The models and the corpora will be distributed online.

\bibliographystyle{ACM-Reference-Format}
\bibliography{my_bib_190624}

\end{document}